\def\year{2019}\relax
\documentclass[letterpaper]{article} 
\usepackage{aaai19}  
\usepackage{times}  
\usepackage{helvet}  
\usepackage{courier}  
\usepackage[hyphens]{url}  
\usepackage{graphicx}  
\usepackage{times}
\usepackage{amsmath}
\usepackage{latexsym}
\usepackage{booktabs}
\usepackage{color}

\usepackage{epsfig}
\usepackage[ruled,vlined]{algorithm2e}
\usepackage{url}
\newcommand{\citet}[1]
  {\citeauthor{#1} ̃\shortcite{#1}}
\newcommand{\citep}{\cite}

\begin{document}
%
\title{Domain Agnostic Real-Valued Specificity Prediction}
\author{Wei-Jen Ko$^*$,
Greg Durrett$^*$,
Junyi Jessy Li$^\dagger$\\
$^*$ Department of Computer Science\\
$^\dagger$ Department of Linguistics\\
The University of Texas at Austin\\
{\tt wjko@cs.utexas.edu},
{\tt gdurrett@cs.utexas.edu}, 
{\tt jessy@austin.utexas.edu}}

\maketitle
\begin{abstract}
Sentence specificity quantifies the level of detail in a sentence, characterizing the organization of information in discourse. While this information is useful for many downstream applications, specificity prediction systems predict very coarse labels (binary or ternary) and are trained on and tailored toward specific domains (e.g., news). The goal of this work is to generalize specificity prediction to domains where no labeled data is available and output more nuanced real-valued specificity ratings.

We present an unsupervised domain adaptation system for sentence specificity prediction, specifically designed to output real-valued estimates from binary training labels. To calibrate the values of these predictions appropriately, we regularize the posterior distribution of the labels towards a reference distribution. We show that our framework generalizes well to three different domains with 50\%-68\% mean absolute error reduction than the current state-of-the-art system trained for news sentence specificity. We also demonstrate the potential of our work in improving the quality and informativeness of dialogue generation systems.
\end{abstract}

\section{Introduction}

The specificity of a sentence measures its ``quality of belonging or relating uniquely to a particular subject''\footnote{Definition from the Oxford Dictionary}~\cite{lugini-litman:2017:BEA}. It is often pragmatically defined as the level of detail in the sentence~\cite{louis-nenkova:2011:IJCNLP-2011,li2015specificity}. When communicating, specificity is adjusted to serve the intentions of the writer or speaker~\cite{grice1975logic}. In the examples below, the second sentence is clearly more specific than the first one:
\begin{quote}
\small
{\bf Ex1:} This brand is very popular and many people use its products regularly. \\
{\bf Ex2:} Mascara is the most commonly worn cosmetic, and women will spend an average of  \$4,000 on it in their lifetimes.
\end{quote}

Studies have demonstrated the important role of sentence specificity in reading comprehension~\cite{dixon1987processing} and in establishing common ground in dialog~\cite{djalali2011modeling}. 
It has also been shown to be a key property in analyses and applications, such as 
summarization~\cite{Louis:2011:TSI:2107679.2107684}, 
argumentation mining~\cite{swanson2015argument}, 
political discourse analysis~\cite{cook2016content}, 
student discussion assessment~\cite{luo2016determining,lugini-litman:2017:BEA}, 
deception detection~\cite{kleinberg2017using}, 
and 
dialogue generation~\cite{zhang2018learning}.

Despite their usefulness, prior sentence specificity predictors~\cite{louis-nenkova:2011:IJCNLP-2011,li2015specificity,lugini-litman:2017:BEA} are trained with sentences from specific domains (news or classroom discussions), 
and have been found to fall short when applied to other domains~\cite{kleinberg2017using,lugini-litman:2017:BEA}. They are also trained to label a sentence as either general or specific~\cite{louis-nenkova:2011:IJCNLP-2011,li2015specificity}, or low/medium/high specificity~\cite{lugini-litman:2017:BEA}, though in practice specificity has been analyzed as a continuous value~\cite{Louis:2011:TSI:2107679.2107684,louis2013corpus,swanson2015argument,cook2016content,luo2016determining,kleinberg2017using}, as it should be \cite{LREC}.

In this work, we present an unsupervised domain adaptation system for sentence specificity prediction, specifically designed to output real-valued estimates. It effectively generalizes sentence specificity analysis to domains where no labeled data is available, and outputs values that are close to the real-world distribution of sentence specificity.

Our main framework is an unsupervised domain adaptation system based on Self-Ensembling~\cite{mt,se} 
that simultaneously reduces source prediction errors and generates feature representations that are robust against noise and across domains. Past applications of this technique have focused on computer vision problems; to make it effective for text processing tasks, we modify the network to better utilize labeled data in the source domain and explore several data augmentation methods for text.

We further propose a posterior regularization technique \cite{ganchev2010posterior} that generally applies to the scenario where it is easy to get coarse-grained categories of labels, but fine-grained predictions are needed. Specifically, our regularization term seeks to move the distribution of the classifier posterior probabilities closer to that of a pre-specified target distribution, which in our case is a specificity distribution derived from the source domain.

Experimental results show that our system generates more accurate real-valued sentence specificity predictions that correlate well with human judgment, across three domains that are vastly different from the source domain (news): Twitter, Yelp reviews and movie reviews. Compared to a state-of-the-art system trained on news data~\cite{li2015specificity}, our best setting achieves a 50\%-68\% reduction in mean absolute error and increases Kendall's Tau and Spearman correlations by 0.07-0.10 and 0.12-0.13, respectively.

Finally, we conduct a task-based evaluation that demonstrates the usefulness of sentence specificity prediction in open-domain dialogue generation. 
Prior work showed that the quality of responses from dialog generation systems can be significantly improved if short examples are removed from training~\cite{dialogue}, potentially preventing the system from overly favoring generic responses~\cite{sordoni2015neural,mou2016sequence}. 
We show that predicted specificity works more effectively than length, and enables the system to generate more diverse and informative responses with better quality.
 
In sum, the paper's contributions are as follows:
\begin{itemize}
\item An unsupervised domain adaptation framework for sentence specificity prediction, available at \\ \url{https://github.com/wjko2/Domain-Agnostic-Sentence-Specificity-Prediction}
\item A regularization method to derive real-valued predictions from training data with binary labels;
\item A task-based evaluation that shows the usefulness of our system in generating better, more informative responses in dialogue.
\end{itemize}

\section{Task setup}

With unsupervised domain adaptation, one has access to labeled sentence specificity in one source domain, and unlabeled sentences in all target domains. The goal is to predict the specificity of target domain data. Our source domain is news, the only domain with publicly available labeled data for training~\cite{louis-nenkova:2011:IJCNLP-2011}. We crowdsource sentence specificity for evaluation for three target domains: Twitter, Yelp reviews and movie reviews. The data is described in Section~\ref{sec:data}.

Existing sentence specificity labels in news are {\em binary}, i.e., a sentence is either general or specific. 
However, in practice, real-valued estimates  of sentence specificity are widely adopted~\cite{Louis:2011:TSI:2107679.2107684,louis2013corpus,swanson2015argument,cook2016content,luo2016determining,kleinberg2017using}. Most of these work directly uses the classifier posterior distributions, although we will later show that such distributions do not follow the true distribution of sentence specificity (see Figure~\ref{fig:p3}, {\em Speciteller} vs.\ {\em real}). 
We aim to produce accurate real-valued specificity estimates despite the binary training data. Specifically, the test sentences have real-valued labels between 0 and 1. We evaluate our system using mean absolute error, Kendall's Tau and Spearman correlation.

\section{Architecture}
Our core technique is the Self-Ensembling~\cite{mt,se} of a single base classification model that utilizes data augmentation, 
with a distribution regularization term to generate accurate real-valued predictions from binary training labels.

\begin{figure}[!t]
  \centering
  \includegraphics[scale=0.55,trim=-10mm 95mm 80mm 25mm]{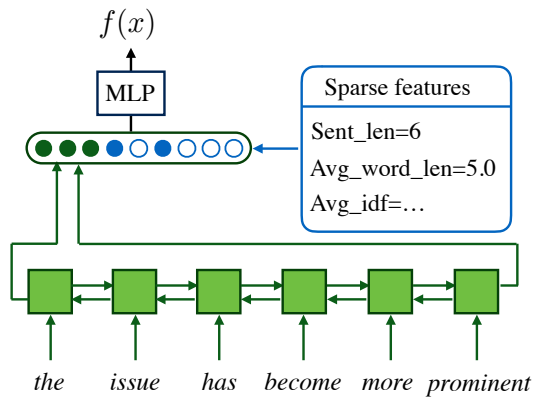}
  \caption{Base model for sentence specificity prediction. The sentence $x$ is encoded with a BiLSTM combined with sparse features, and fed to a MLP to predict specificity $f(x)$.}
  \label{fig:p}
\end{figure}

\begin{figure*}[!t]
  \centering
  \includegraphics[width=0.9	\textwidth]{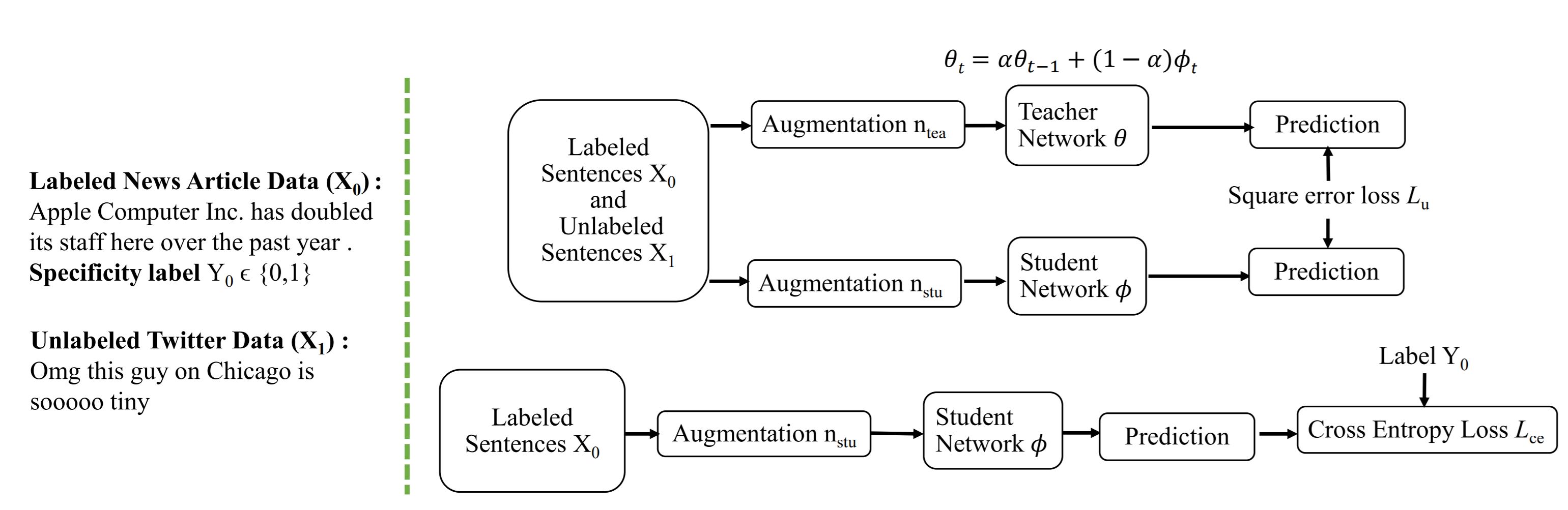}
  \caption{Our unsupervised domain adaptation network, showing the consistency component for representation learning (above) and the prediction component (below). We also show examples of a labeled source sentence and an unlabeled target sentence.}
  \label{fig:p2}
\end{figure*}

\subsection{Base model}\label{sec:base}
Figure~\ref{fig:p} depicts the base model for sentence specificity prediction. The overall structure is similar to that in \citet{lugini-litman:2017:BEA}. Each word in the sentence is encoded into an embedding vector and the sentence is passed through a bidirectional Long-Short Term Memory Network (LSTM)~\cite{Hochreiter:1997:LSM:1246443.1246450} to generate a representation of the sentence. This representation is concatenated with a series of hand crafted features, and then passed through a multilayer perceptron to generate specificity predictions $f(x) \in [0,1]$. Training treats these as probabilities of the positive class, but we will show in Section~\ref{sec:regularization} how they can be adapted to make real-valued predictions. 

Our hand crafted features are taken from \citet{li2015specificity}, 
including: the number of tokens in the sentence; the number of numbers, capital letters and punctuations in the sentence, normalized by sentence length; the average number of characters in each word; the fraction of stop words; the number of words that can serve as explicit discourse connectives; the fraction of words that have sentiment polarity or are strongly subjective; the average familiarity and imageability of words; and the minimum, maximum, and average inverse document frequency (idf) over the words in the sentence.\footnote{We use existing idf values provided by \cite{li2015specificity} calculated from the New York Times corpus~\cite{sandhaus08}.} In preliminary experiments, we found that combining hand-crafted features with the BiLSTM performed better than using either one individually.

\subsection{Unsupervised domain adaptation}

Our unsupervised domain adaptation framework is based on Self-Ensembling~\cite{se}; we lay out the core ideas first and then discuss modifications to make the framework suitable for textual data. 

There are two ideas acting as the driving force behind Self-Ensembling.  
First, adding noise to each data point $x$ can help regularize the model by encouraging the model prediction to stay the same regardless of the noise, creating a {\em manifold} around the data point where predictions are invariant~\cite{rasmus2015semi}. This can be achieved by minimizing a {\em consistency loss} $L_{u}$ between the predictions. 
Second, temporal ensembling---ensemble the same model trained at different time steps---is shown to be beneficial to prediction especially in semi-supervised cases~\cite{laine2016temporal}; in particular, we average model parameters from each time step~\cite{mt}.

These two ideas are realized by using a student network and a teacher network. The parameters of the teacher network are the exponential average of that of the student network, making the teacher a temporal ensemble of the student. Distinct noise augmentations are applied to the input of each network, hence the consistency loss $L_{u}$ is applied between student and teacher predictions. The student learns from labeled source data and minimizes the supervised cross-entropy loss $L_{ce}$. Domain adaptation is achieved by minimizing the consistency loss between the two networks, which can be done with unlabeled target data. The overall loss function is the weighted sum of $L_{ce}$ and $L_{u}$. Figure~\ref{fig:p2} depicts the process.

Concretely, the student and teacher networks are of identical {\em structure} following the base model (Section~\ref{sec:base}), but features distinct {\em noise augmentation}. 
The student network learns to predict sentence specificity from labeled source domain data. The input sentences are augmented with noise ${n}_{stu}$. 
The teacher network predicts the specificity of each sentence with a different noise augmentation ${n}_{tea}$. 
Parameters of the teacher network $\theta$ are updated each time step to be the exponential moving average of the corresponding parameters in the student network. 
The teacher parameter $\theta_t$ at each time step $t$ is
\begin{equation}
{\theta}_{t}={\alpha}{\theta}_{t-1}+(1-\alpha){\phi}_{t}
\end{equation}
where $\alpha$ is the degree of weighting decay, a constant between 0 and 1. $\phi$ denotes parameters for the student network.

The consistency loss is defined as the squared difference between the predictions of the student and teacher networks~\cite{mt}:
\begin{equation}
L_u=(f({n}_{stu}(x)|\phi)-f({n}_{tea}(x)|\theta))^2
\end{equation}
where $f$ denotes the base network and x denotes the input sentence. The teacher network is not involved when minimizing the supervised loss $L_{ce}$.

An important difference between our work and \citet{se} is that in their work only unlabeled target data contributes to the consistency loss ${L}_{u}$. Instead, we use both source and target sentences, bringing the predictions of the two networks close to each other not only on the target domain, but also the source domain. Unlike many vision tasks where predictions intuitively stay the same with different types of image augmentation (e.g., transformation and scaling), text is more sensitive to noise. Self-Ensembling relies on heavy augmentation on both student and teacher networks, and our experiments revealed that incorporating source data in the consistency loss term mitigates additional biases from noise augmentation. 

At training time, the teacher network's parameters are fixed during gradient descent, and the gradient only propagates through the student network. After each update of the student network, we recalculate the weights of the teacher network using exponential moving average. 
At testing time, we use the teacher network for prediction.

\paragraph{Noise augmentation} An important factor contributing to the effectiveness of  Self-Ensembling is applying noise to the inputs to make them more robust against domain shifts. For computer vision tasks, augmentation techniques including affine transformation, scaling, flipping and cropping could be used~\cite{se}. However these operations could not be used on text. We designed several noise augmentations for sentences, including: adding Gaussian noise to both the word embeddings and shallow features; randomly removing words in a sentence; substituting word embeddings with a random vector, or a zero vector like applying dropout. To produce enough variations of data, augmentation is applied to $\sim$half of the words in a sentence.

\subsection{Regularizing the posterior distribution}\label{sec:regularization}

Sentence specificity prediction is a task where the existing training data have binary labels, while real-valued outputs are desirable. Prior work has directly used classifier posterior probabilities. However, the posterior distribution and the true specificity distribution are quite different (see Figure~\ref{fig:p3}, {\em Speciteller} vs.\ {\em real}). 
We propose a regularization term to bridge the gap between the two.

Specifically, we view the posterior probability distribution as a latent distribution, which allows us to apply a variant of posterior regularization~\cite{ganchev2010posterior}, previously used to apply pre-specified constraints to latent variables in structured prediction. Here, we apply a distance penalty between the latent distribution and a pre-specified reference distribution (which, in our work, is from the source domain). \citet{LREC} found that in news, the distribution of sentence specificity is bell shaped, similar to that of a Gaussian. Our analysis on sentence specificity for three target domains yields the same insights (Figure~\ref{fig:p4}).  
We explored two regularization formulations, with and without assuming that the two distributions are Gaussian. Both were successful and achieved similar performance.

Let ${\mu}_p$ and ${\sigma}_p$ be the mean and standard deviation of the predictions (posterior probabilities) in a batch. 
The first formulation assumes that the predictions and reference distributions are Gaussian. It uses the KL divergence between the predicted distribution $p(x)=N({\mu}_p,{\sigma}_p)$ and the reference Gaussian distribution $r(x)=N({\mu}_r,{\sigma}_r)$. The distribution regularization loss can be written as: 
\begin{equation}\label{eq:kl}
{L}_{d} =KL(r||p) =\log\frac{\sigma_p}{\sigma_r}+\frac{\sigma_r^2+(\mu_r-\mu_p)^2}{2\sigma_p^2}-\frac{1}{2}
\end{equation}

The second formulation does not assume Gaussian distributions and only compares the mean and standard deviation of the two distributions using a weighting term $\beta$:
\begin{equation}\label{eq:meanstd}
{L}_{d}=|{\sigma }_{r}-{\sigma }_{p}| +\beta|{\mu }_{r}-{\mu }_{p} |.
\end{equation}

Combining the regularization term $L_d$ into a single objective, the total loss is thus:
\begin{equation}
L={L}_{ce}+{c}_{1}{L}_{u}+{c}_{2}{L}_{d}
\end{equation}
where ${L}_{ce}$ is the cross entropy loss for the source domain predictions, ${L}_{u}$ is the consistency loss. $c_1$ and $c_2$ are weighting hyperparameters.

In practice, this regularization term serves a second purpose. After adding the consistency loss $L_u$, we observed that the predictions are mostly close to each other with values between 0.4 and 0.6, and their distribution resembles a Gaussian with very small variance (c.f.\ Figure \ref{fig:p3} line {\em SE+A}). 
This might be due to the consistency loss pulling all the predictions together, since when all predictions are identical, the loss term will be zero. This regularization can be used to counter this effect, and avoid the condensation of predicted values. 
Finally, this regularization is distinct from class imbalance loss terms such as that used in \citet{se}, which we found early on to hurt performance rather than helping.

\section{Datasets for sentence specificity} \label{sec:data}

\subsection{Source domain} 

The source domain for sentence specificity is news, for which we use three publicly available labeled datasets: (1) training sentences from~\citet{louis-nenkova:2011:IJCNLP-2011} and ~\citet{li2015specificity}, which consists of 1.4K general and 1.4K specific sentences from the Wall Street Journal.  
(2) 900 news sentences crowdsourced for binary general/specific labels~\cite{louis2012corpus}; 
55\% of them are specific. 
(3) 543 news sentences from \citet{LREC}. These sentences are rated on a scale of $0-6$, so for consistency with the rest of the training labels, we pick sentences with average rating  $>3.5$ as general examples and those with average rating $<2.5$ as specific. In total, we have 4.3K sentences with binary labels in the source domain.

\begin{figure}[!t]
  \centering
  \includegraphics[width=0.35\textwidth]{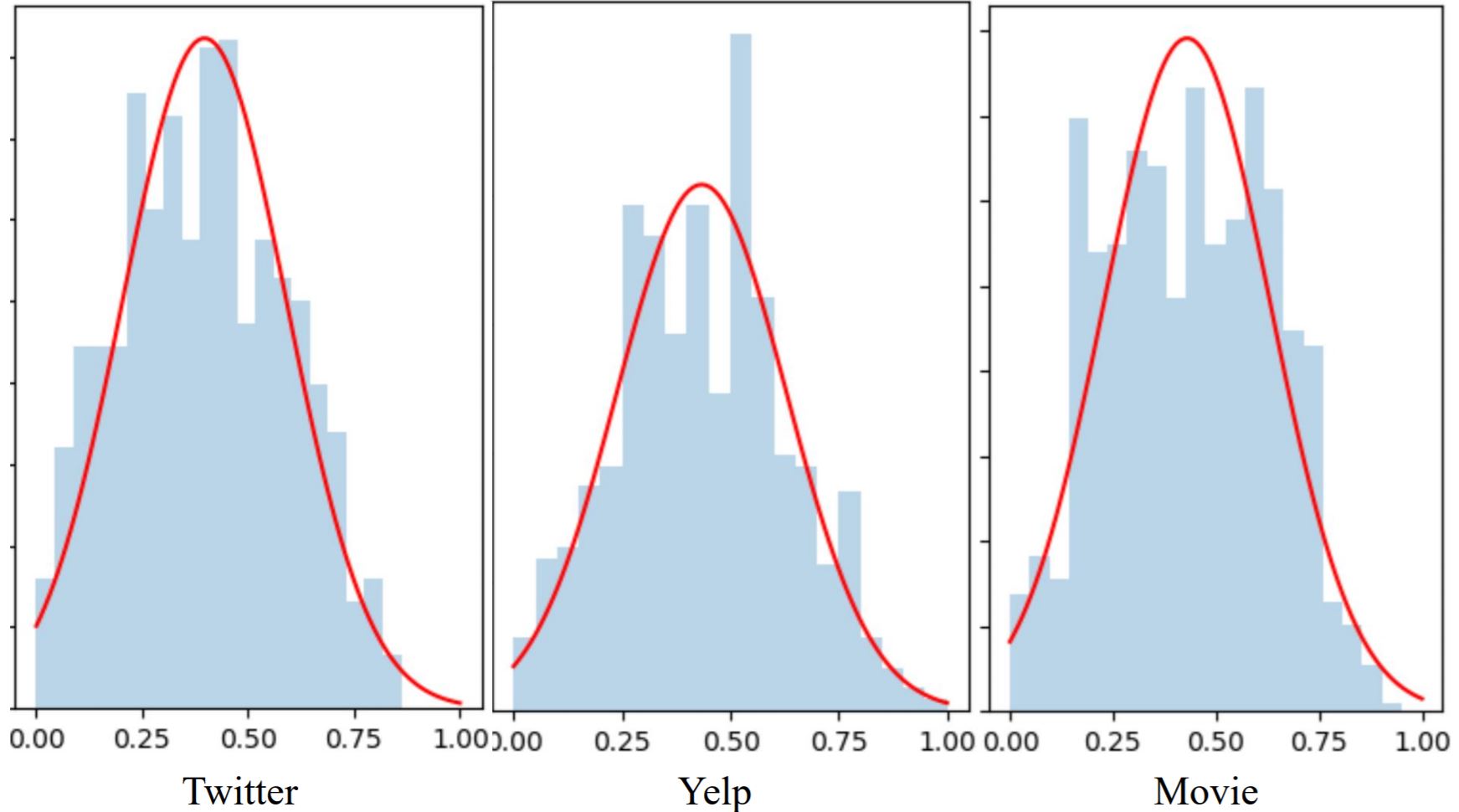}
  \caption{Histograms of specificity distribution for each target domain, shown with a fitted Gaussian distribution.}
  \label{fig:p4}
\end{figure}

\subsection{Target domains}

We evaluate on three target domains: Twitter, Yelp and movie reviews. Since no annotated data exist for these domains, we crowdsource specificity for sentences sampled from each domain using Amazon Mechanical Turk.

We follow the context-independent annotation instructions from~\citet{LREC}. Initially, 9 workers labeled specificity for 1000 sentences in each domain on a scale of 1 (very general) to 5 (very specific), which we rescaled to 0, 0.25, 0.5, 0.75, and 1. Inter-annotator agreement (IAA) is calculated using average Cronbach's alpha~\cite{cronbach} values for each worker. For quality control, we exclude workers with IAA below 0.3, and include the remaining sentences that have at least 5 raters. Our final IAA values fall between 0.68-0.70, compatible with the reported 0.72 from expert annotators in~\citet{LREC}. The final specificity value is aggregated to be the average of the rescaled ratings. 
We also use large sets of unlabeled data in each domain:
\begin{itemize}
\item Twitter: 984 tweets annotated, 50K unlabeled, sampled from ~\citet{preoctiuc2017beyond}.
\item Yelp: 845 sentences annotated, 95K unlabeled, sampled from the Yelp Dataset Challenge 2015~\cite{yelp}. 
\item Movie: 920 sentences annotated, 12K unlabeled, sampled from~\citet{movie}.
\end{itemize}

Figure~\ref{fig:p4} shows the distribution of ratings for the annotated data. We also plot a fitted Gaussian distribution for comparison. Clearly, most sentences have mid-range specificity values, consistent with news sentences~\cite{LREC}. Interestingly the mean and variance for the three distributions are similar to each other and to those from~\citet{LREC}, as show in Table~\ref{tab:specdistmeanstd}. Therefore, we use the source distribution (news, \citet{LREC}) as the reference distribution for posterior distribution regularization, and set ${\mu}_r,{\sigma}_r$ to 0.417 and 0.227 accordingly.

\begin{table}[t]
\begin{center}
\begin{tabular}{ c|c c } 

domain & mean & std.dev  \\
\toprule
Twitter & 0.405& 0.193 \\
Yelp & 0.419& 0.198  \\
Movie  & 0.426 & 0.206 \\
News~\cite{LREC} & 0.417 & 0.227  \\
\bottomrule
\end{tabular}
\end{center}
\caption{Mean and standard deviation of sentence specificity for each domain.}
\label{tab:specdistmeanstd}
\end{table}

\section{Experiments}

We now evaluate our framework on predicting sentence specificity for the three target domains. We report experimental results on a series of settings to evaluate the performance of different components. 

\subsection{Systems}

\noindent{\bf Length baseline} This simple baseline predicts specificity proportionally to the number of words in the sentence. Shorter sentences are predicted as more general and longer sentences are predicted as more specific.

\noindent{\bf Speciteller baseline}
Speciteller~\cite{li2015specificity} is a semi-supervised system trained on news data with binary labels. The posterior probabilities of the classifier are used directly as specificity values.

\noindent{\bf Self-ensembling baseline (SE)} 
Our system with the teacher network, but only using exponential moving average (without the consistency loss or distribution regularization).

\noindent{\bf Distribution only (SE+D)} 
Our system with distribution regularization ${L}_{d}$ using the mean and standard deviation (Eq.~\ref{eq:meanstd}), but without the consistency loss ${L}_{u}$.

\noindent{\bf Adaptation only (SE+A)} 
Our system with the consistency loss ${L}_{u}$, but without distribution regularization ${L}_{d}$.

\noindent{\bf SE+AD (KL)} 
Our system with both ${L}_{u}$ and ${L}_{d}$ using KL divergence (Eq.~\ref{eq:kl}).	

\noindent{\bf SE+AD (mean-std)} 
Our system with both ${L}_{u}$ and ${L}_{d}$ using mean and standard deviation (Eq.~\ref{eq:meanstd}). 	

\noindent{\bf SE+AD (no augmentation)}
We also show the importance of noise augmentation, by benchmarking the same setup as SE+AD (mean-std) without data augmentation.

\begin{table*}[t]
\begin{center}
\small
\begin{tabular}{ c| c c|c c c| c c |c } 
& \multicolumn{2}{c|}{Baselines} & & & & \multicolumn{3}{c}{SE+AD}\\ \hline
metric & Length & Speciteller & SE & SE+D  & SE+A & mean-std & KL & no aug.\\
\toprule
\multicolumn{9}{c}{Twitter}\\ \midrule
Spearman & 0.445 & 0.553 & 0.622$\pm$0.028 & 0.610$\pm$0.011& 0.670$\pm$0.004 &0.676$\pm$0.004&\textbf{0.679$\pm$0.005}&0.666 \\
Kendall's Tau  & 0.324 & 0.413 & 0.437$\pm$0.024 & 0.427$\pm$0.008& 0.480$\pm$0.004 &\textbf{0.487$\pm$0.005}&0.482$\pm$0.006&0.480 \\
MAE & - & 0.237 & 0.148$\pm$0.006 & 0.125$\pm$0.001& 0.151$\pm$0.005 &\textbf{0.113$\pm$0.001}&0.115$\pm$0.002&0.115 \\
\midrule
\multicolumn{9}{c}{Yelp}\\ \midrule
Spearman & 0.676 & 0.633 & 0.731$\pm$0.009 & 0.721$\pm$0.009& 0.735$\pm$0.001 &\textbf{0.750$\pm$0.016}&0.743$\pm$0.010&0.728 \\
Kendall's Tau  & 0.522 & 0.481 & 0.548$\pm$0.006 & 0.536$\pm$0.008& 0.544$\pm$0.001 &\textbf{0.555$\pm$0.010}&0.546$\pm$0.014&0.533 \\
MAE & - & 0.325 & 0.165$\pm$0.016 & 0.120$\pm$0.010& 0.137$\pm$0.005 &0.107$\pm$0.003&\textbf{0.105$\pm$0.002}&0.109 \\
\midrule
\multicolumn{9}{c}{Movie}\\ \midrule
Spearman & 0.581 & 0.575 & 0.684$\pm$0.004 & 0.664$\pm$0.004& 0.680$\pm$0.007 &0.702$\pm$0.003&\textbf{0.706$\pm$0.030}&0.669 \\
Kendall's Tau  & 0.435 & 0.418 & 0.498$\pm$0.004 & 0.487$\pm$0.010& 0.502$\pm$0.010 &0.519$\pm$0.015&\textbf{0.522$\pm$0.024}& 0.484\\
MAE & - & 0.226 & 0.143$\pm$0.009 & 0.124$\pm$0.009& 0.148$\pm$0.001 &\textbf{0.114$\pm$0.001}&\textbf{0.114$\pm$0.006} &0.118\\
\bottomrule
\end{tabular}
\end{center}
\caption{Sentence specificity prediction results (across 3 runs) for: Length and Speciteller baselines, Self-Ensembling baseline (SE), SE with mean-std distribution regularization (SE+D), with consistency loss (SE+A), and both (SE+AD). Also showing SE+AD without data augmentation (no aug.).}
\label{tab:results}
\end{table*}

\subsection{Training details}

Hyperparameters are tuned on a validation set of 200 tweets that doesn't overlap with the test set. We then use this set of parameters for all testing domains. 
The LSTM encoder generates 100-dimensional representations. For the multilayer perceptron, we use 3 fully connected 100-dimensional layers. We use ReLU activation  with batch normalization. For the Gaussian noise in data augmentation, we use standard deviation 0.1 for word embeddings and 0.2 for shallow features. The probabilities of deleting a word and replacing a word vector are 0.15. The exponential moving average decay $\alpha$ is 0.999. Dropout rate is 0.5 for all layers. The batch size is 32. ${c}_{1}=1000$, ${c}_{2}=10$ for KL loss and 100 for mean and std.dev loss. $\beta=1$. We fix the number of training to be 30 epochs for SE+A and SE+AD, 10 epochs for SE, and 15 epochs for SE+D. We use the Adam optimizer with learning rate 0.0001, ${\beta}_{1}=0.9$, ${\beta}_{2}=0.999$. As discussed in Section~\ref{sec:data}, posterior distribution regularization parameters ${\mu}_r,{\sigma}_r$ are set to be those from \citet{LREC}.

\subsection{Evaluation metrics}
We use 3 metrics to evaluate real-valued predictions: (1) the {\bf Spearman correlation} between the labeled and predicted specificity values, higher is better; (2) the pairwise {\bf Kendall's Tau} correlation, higher is better; (3) {\bf mean absolute error (MAE)}: $\sum|Y-X|/n$, lower is better.

\subsection{Results and analysis}
\begin{figure}[!t]
  \centering
  \includegraphics[width=0.4\textwidth]{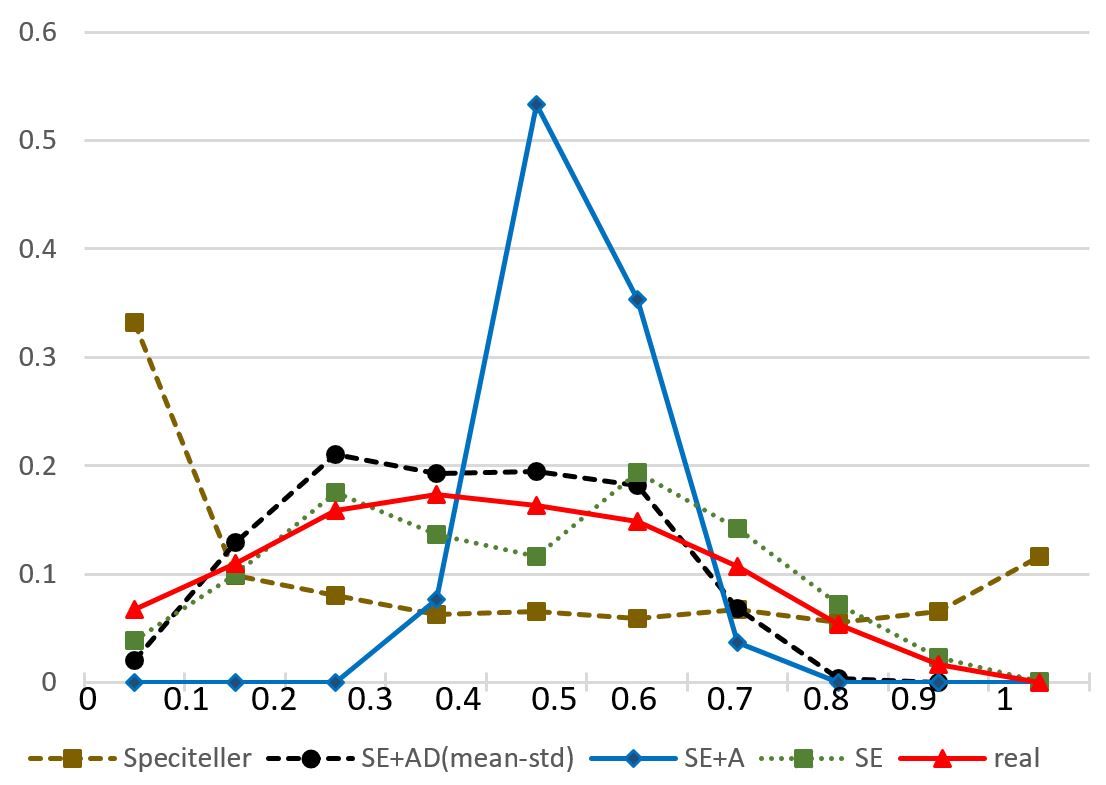}
  \caption{Distribution of predictions for: Speciteller, Self-Ensembling baseline (SE), SE with consistency loss (SE+A), and also with distribution regularization (SE+AD).}
  \label{fig:p3}
\end{figure}

Table~\ref{tab:results} shows the full results for the baselines and each configuration of our framework. For analysis, we also plot in Figure~\ref{fig:p3} the true specificity distributions in Twitter test set, predicted distributions for Speciteller, the Self-Ensembling baseline (SE), SE with adaptation (SE+A) and also with distribution regularization (SE+AD).

Speciteller, which is trained on news sentences, cannot generalize well to other domains, as it performs worse than just using sentence length for two of the three domains (Yelp and Movie). From Figure~\ref{fig:p3}, we can see that the prediction mass of Speciteller is near the extrema values 0 and 1, and the rest of the predictions falls uniformly in between. These findings confirm the need of a  generalizable system.

Across all domains and all metrics, the best performing system is our full system with domain adaptation and distribution regularization (SE+AD with mean-std or KL), showing that the system generalizes well across different domains. Using a paired Wilcoxon test, it significantly ($p<0.001$) outperforms Speciteller in terms of MAE; it also achieved higher Spearman and Kendall's Tau correlations than both Length and Speciteller. 

Component-wise, the Self-Ensembling baseline (SE) achieves significantly lower MAE than Speciteller, and higher correlations than either baseline. Figure~\ref{fig:p3} shows that unlike Speciteller, the SE baseline does not have most of its prediction mass near 0 and 1, demonstrating the effectiveness of temporal ensembling. 
Using both the consistency loss ${L}_{u}$ and  the distribution regularization $L_d$ achieves the best results on all three domains; however 
adding only ${L}_{u}$ (SE+A) or $L_d$ (SE+D) improves for some measures or domains but not all. 
This shows that both terms are crucial to make the system robust across domains. 

The improvements from distribution regularization are visualized in Figure~\ref{fig:p3}. With SE+A, most of the predicted labels are between 0.4 and 0.6. Applying distribution regularization (SE+AD) makes them much closer to the real distribution of specificity. 
With respect to the two formulations of regularization (KL and mean-std), both are effective in generating more accurate real-valued estimates. Their performances are comparable, hence using only the mean and standard deviation values, without explicitly modeling the reference Gaussian distribution, works equally well.  

Finally, without data augmentation (column no aug.), the correlations are clearly lower than our full model, stressing the importance of data augmentation in our framework.

\section{Specificity in dialogue generation}

We also evaluate our framework in open-domain dialogue. This experiment also presents a case for the usefulness of an effective sentence specificity system in dialogue generation.

With {\sc seq2seq} dialogue generation models, \cite{dialogue} observed significant quality improvement by removing training examples with short responses during preprocessing; this is potentially related to this type of models favor generating non-informative, generic responses~\cite{sordoni2015neural,mou2016sequence}. We show that filtering training data by predicted specificity results in responses of higher quality and informativeness than filtering by length.

\subsection{Task and settings}
We implemented a {\sc seq2seq} question-answering bot with attention using OpenNMT\cite{OpenNMT}. The bot is trained on  OpenSubtitles~\cite{opensubtitles} following prior work~\cite{dialogue}. We restrict the instances to question-answer pairs by selecting consecutive sentences with the first sentence ending with question mark, the second sentence without question mark and follows the first sentence by less than 20 seconds, resulting in a 14M subcorpus.

The model uses two hidden layers of size 2048, optimized with Adam with learning rate 0.001. The batch size is 64. While decoding, we use beam size 5; we block repeating n-grams, and constrain the minimum prediction length to be 5. These parameters are tuned on a development set. 
We compare two ways to filter training data during preprocessing:

\noindent {\bf Remove Short:} Following \citet{dialogue}, remove training examples with length of responses shorter than a threshold of 5. About half of the data are removed.

\noindent {\bf Remove General:} Remove predicted general responses from training examples using our system. We use the responses in the OpenSubtitles training set as the unlabeled target domain data during training. We remove the least specific responses such that the resulting number of examples is the same as Remove Short.

For fair comparison, at test time, we adjust the length penalty described in \citet{google} for both models, so the average response lengths are the same among both models.

\begin{table}[t]
\centering
\small 
\begin{tabular}{ c|c c }
& Remove Short&  Remove General\\
\toprule
unigram diversity & 0.0177& \textbf{0.0199}\\

bigram diversity & 0.0833&\textbf{ 0.0977}\\

perplexity& 51.97 &\textbf{46.92}\\
\bottomrule
\end{tabular}
\caption{Perplexity and diversity scores for Remove Short and Remove General.}
\label{tab:dialogue3}
\end{table}

\begin{table}[t]
\centering
\small
\begin{tabular}{ c|c c c } 
 & Remove&  Remove& \\
 & Short wins & General wins & tie \\\toprule
informativeness &29.0& 38.9  &32.1  \\
quality &  28.9&34.7  &36.4  \\
\bottomrule
\end{tabular}
\caption{Human evaluation results for Remove Short and Remove General.}
\label{tab:dialogue1}
\end{table}

\subsection{Evaluation}
We use automatic measures and human evaluation as in \citet{diversity} and \citet{dialogue}.

Table~\ref{tab:dialogue3} shows the diversity and perplexity of the responses. Diversity is calculated as the type-token ratio of unigrams and bigrams. The test set for these two metrics is a random sample of 10K instances from OpenSubtitles that doesn't overlap with the training set. Clearly, filtering training data according to specificity results in more diverse responses with lower perplexity than filtering by length.

We also crowdsource human evaluation for quality; in addition, we evaluate the systems for response informativeness. Note that in our instructions, informativeness means the {\em usefulness} of information and is a distinct measure from specificity. The original training data of specificity are linguistically annotated and involves only the change in the level of details~\cite{louis-nenkova:2011:IJCNLP-2011}. Separate experiments are conducted to avoid priming. We use a test set of 388 instances, including questions randomly sampled from OpenSubtitles that doesn't overlap with the training set, and 188 example questions from previous dialogue generation papers, including \citet{conversation}. We use Amazon MechenicalTurk for crowdsourcing. 7 workers chose between the two responses to the same question.  

Table  \ref{tab:dialogue1} shows the human evaluation comparing Remove Short vs.\ Remove General.
Removing predicted general responses performs better than removing short sentences, on both informativeness and quality and on both test sets. 
This shows that sentence specificity is a superior measure for training data preprocessing than sentence length.

\section{Related Work}

Sentence specificity prediction as a task is proposed by~\citet{louis-nenkova:2011:IJCNLP-2011}, who repurposed discourse relation annotations from WSJ articles~\cite{Prasad08thepenn} for sentence specificity training. \citet{li2015specificity} incorporated more news sentences as unlabeled data. \citet{lugini-litman:2017:BEA} developed a system to predict sentence specificity for classroom discussions, however the data is not publicly available. All these systems are classifiers trained with categorical data (2 or 3 classes).

We use Self-Ensembling~\cite{se} as our underlying framework. Self-Ensembling builds on top of Temporal Ensembling~\cite{laine2016temporal} and the Mean-Teacher network~\cite{mt}, both of which originally proposed for semi-supervised learning. In visual domain adaptation, Self-Ensembling shows superior performance than many recently proposed approaches \cite{RG,DCRN,SBDA,ADA,ADDA,GTA}, including GAN-based approaches. To the best of our knowledge, this approach has not been used on language data.

\section{Conclusion}
We present a new model for predicting sentence specificity. We augment the Self-Ensembling method~\cite{se} for unsupervised domain adaptation on text data. We also regularize the distribution of predictions to match a reference distribution. Using only binary labeled sentences from news articles as source domain, our system could generate real-valued specificity predictions on different target domains, significantly outperforming previous work on sentence specificity prediction. Finally, we show that sentence specificity prediction can potentially be beneficial in improving the quality and informativeness of dialogue generation systems.

\section*{Acknowledgments}
This research was partly supported by the Amazon Alexa Graduate Fellowship. We thank the anonymous reviewers for their helpful comments.

\bibliography{references}
\bibliographystyle{aaai}

\end{document}